\documentclass[runningheads]{llncs}
\usepackage{eccv}
\usepackage{eccvabbrv}
\usepackage{xcolor}
\usepackage{multirow}
\usepackage{booktabs}
\usepackage{pifont}
\usepackage{graphicx}
\usepackage{subcaption} 
\usepackage{adjustbox}
\usepackage{amsmath}
\usepackage{caption}
\usepackage[accsupp]{axessibility}  
\usepackage{hyperref}
\usepackage{wrapfig}
\usepackage{orcidlink}
\begin{document}
\title{Benchmarking Dynamic Affective Reasoning: A Viewer-Centric Video Emotion Dataset}
\titlerunning{Dynamic Affective Reasoning}
\author{
Zhiyan Zhang\inst{1}\orcidlink{0009-0009-0992-5900} \and
Peipei Song\inst{1}\orcidlink{0000-0001-6764-3375}\thanks{Corresponding author.} \and
Jinpeng Hu\inst{2}\orcidlink{0000-0003-4090-7494} \and
Jingyang Jia\inst{1}\orcidlink{0009-0003-5806-5659} \and
Xun Yang\inst{1}\orcidlink{0000-0003-0201-1638} \and
Xiaojun Chang\inst{1}\orcidlink{0000-0002-7778-8807}
}

\authorrunning{Z. Zhang et al.}

\institute{
University of Science and Technology of China, Hefei, China\\
\email{
zzyhang02@gmail.com,
beta.songpp@gmail.com,
jjygood@mail.ustc.edu.cn,
xyang21@ustc.edu.cn,
xjchang@ustc.edu.cn
}
\and
Hefei University of Technology, Hefei, China\\
\email{135858hjp@gmail.com}
}

\maketitle

\begin{abstract}
Video emotion analysis is typically framed as a static classification problem, treating each clip as an independent labeled unit. However, such a formulation overlooks a key psychological fact: emotions change as a result of cumulative reactions to consecutive causal events. To bridge this gap, we introduce \textbf{DAR} (Dynamic Affective Reasoning),  the first large-scale benchmark for viewer-centric affect transitions and causal reasoning over consecutive video events. DAR contains 15,087 videos and 36,908 event-aligned affective segments annotated with 27 emotion categories. Unlike existing video-based emotion datasets, DAR presents a new viewer-centric perspective on fine-grained emotional expressions and transitions, and provides dense, temporally grounded, and causally explicit reasoning chains. Based on DAR, we formally define three challenging tasks: affective segmentation, fine-grained emotion classification, and affective reasoning. 
Complementing this benchmark, we propose \textbf{DAR-R1}, a two-stage framework that combines supervised fine-tuning with Group Relative Policy Optimization. Experiments across 10+ MLLMs show that DAR-R1 sets a new state-of-the-art for dynamic affective reasoning, in terms of both emotional localization and affective reasoning. Project page: https://github.com/Zhang-Zhiyan/DAR.
\end{abstract}

\section{Introduction}
\label{sec:intro}

\begin{figure}[t]
    \centering
    \includegraphics[width=\linewidth]{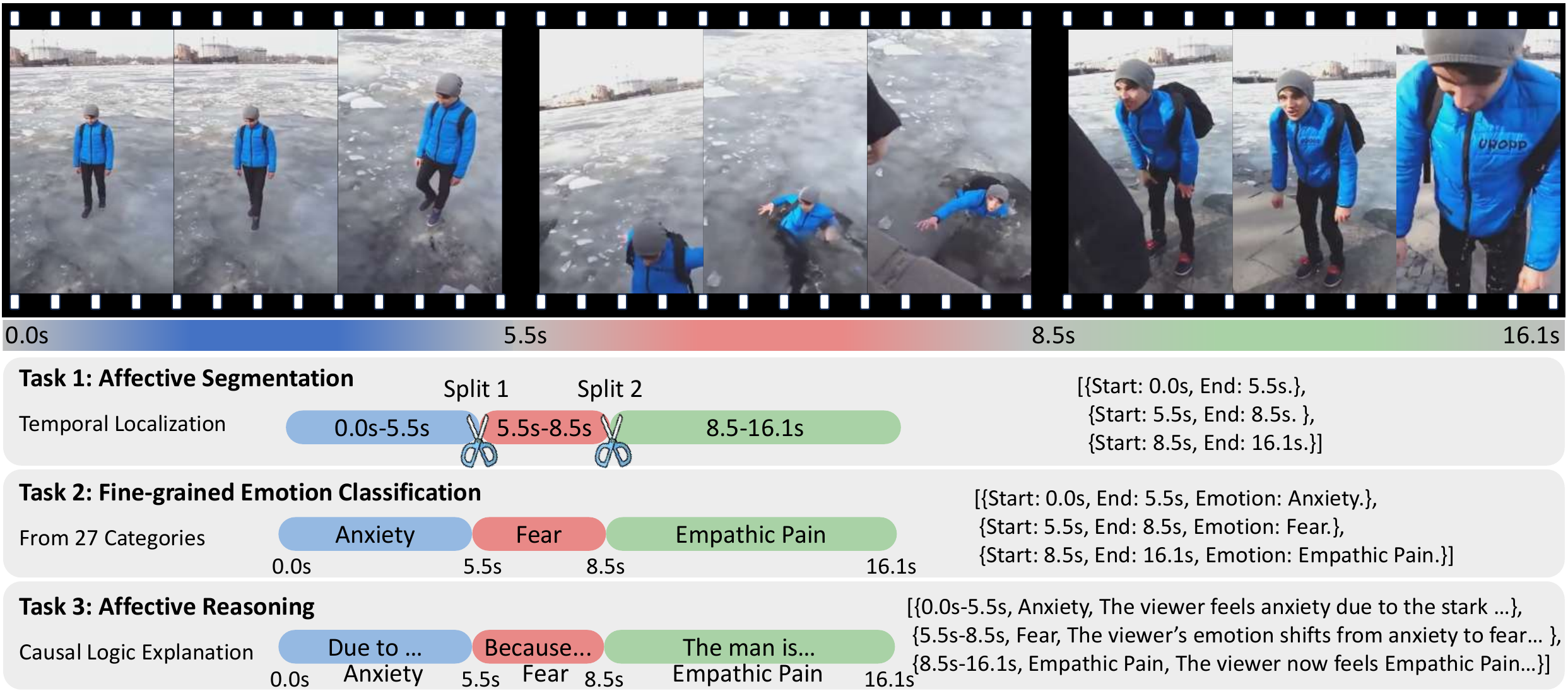}
    \caption{\textbf{Overview of the DAR Benchmark Tasks.} We formulate three hierarchical tasks: (1) \textit{Affective Segmentation} for locating temporal boundaries of emotion shifts; (2) \textit{Fine-grained Emotion Classification} utilizing a 27-category viewer-centric taxonomy; and (3) \textit{Affective Reasoning} for generating causal explanations grounded in visual evidence.}
    \label{fig:task_overview}
    \vspace{-10pt}
\end{figure}

Affective computing~\cite{pantic2005affective} aims to equip machines with the ability to perceive, interpret, and reason about human affect, enabling applications from human--computer interaction~\cite{filippini2020thermal} to psychological counseling and psychotherapy~\cite{liu2021towards,yannakakis2018enhancing,xu2025multiagentesc,xu2025prompt}. As AI systems become embedded in everyday products~\cite{yadegaridehkordi2019affective}, robust emotion understanding is increasingly essential for natural, adaptive, and socially aware interaction~\cite{wang2025dual}. Recent Multimodal Large Language Models (MLLMs)~\cite{yang2025qwen3,chen2024internvl,achiam2023gpt,yang2026fine} have further advanced video-based emotion understanding and recognition~\cite{han2025benchmarking,cheng2024emotion}. Foundational benchmarks such as DFEW~\cite{jiang2020dfew} and VCE~\cite{mazeika2022would} enabled systematic evaluation of video emotion understanding, while Social-IQ~\cite{zadeh2019social} extended evaluation to broader social and contextual cues. More recently, MER~\cite{lian2023mer,cheng2024emotion} incorporated evidence-based reasoning into emotion recognition.

However, key aspects of affect remain underexplored in current benchmarks. Prior works either emphasize describing visual content~\cite{maaz2024video,su2023pandagpt,lin2024video,zhang2026stimuli} or improve character-centric emotion recognition~\cite{lian2025affectgpt,zhang2025videmo}, but rarely model how consecutive events drive moment-to-moment shifts in a viewer's emotional state. Furthermore, most datasets adopt coarse taxonomies following Ekman's theory~\cite{ekman}, which are often insufficient to represent the diversity of real-world emotion reactions. As a result, video emotion analysis is still largely treated as assigning a single label to an entire clip~\cite{lian2023mer,hu2025beyond} or focusing primarily on character emotions~\cite{lian2025affectgpt}, without explicit boundaries for emotion shifts or explanations of why they occur. These omissions hinder the deployment of emotionally intelligent agents, which are expected to respond continuously and causally to users. Therefore, we advocate reframing the video emotion analysis problem as fine-grained dynamic affective reasoning.

To support this paradigm shift, we draw inspiration from the \textit{Affective Events Theory} (AET)~\cite{weiss1996affective}, which views emotions as reactions to consecutive events that accumulate over time: a viewer's current affect is shaped by preceding stimuli. Guided by AET, we introduce \textbf{DAR}, a large-scale viewer-centric video emotion benchmark, designed to support multi-level emotion cognitive tasks. As illustrated in Figure~\ref{fig:task_overview}, we define three challenging tasks based on DAR: (1) Affective Segmentation, locating temporal boundaries of emotional shifts; (2) Fine-grained Emotion Classification, identifying specific emotional states from 27 categories; and (3) Affective Reasoning, explaining the causal logic behind each emotion. To scale annotation while preserving temporal precision, we adopt a coarse-to-fine data construction pipeline that couples high-level semantic boundary proposals with low-level visual cut refinement.

Extensive evaluations on 10+ mainstream MLLMs reveal that neither general-purpose models~\cite{maaz2024video,yang2025qwen3} nor existing emotion-specialized models~\cite{lian2025affectgpt,zhang2025videmo} satisfy the requirements of dynamic affective reasoning. Generalist models often struggle with strict temporal schemas and boundary localization, whereas emotion-focused models typically lack the generative capacity to produce temporally grounded, viewer-centric explanations. To provide a strong baseline, we propose \textbf{DAR-R1}, a two-stage tuning pipeline consisting of cold-start supervised training for structural adaptation followed by GRPO-based reinforcement learning~\cite{guo2025deepseek}. By explicitly rewarding boundary precision and reasoning quality, \textsc{DAR-R1} better aligns MLLMs with fine-grained affect dynamics, yielding consistent gains over state-of-the-art baselines on \textsc{DAR}.

In summary, our contributions are fourfold:
\begin{enumerate}
    \item \textit{\textbf{New Task:}} We introduce a task of \emph{Dynamic Affective Reasoning}, reframing video emotion analysis from static classification to temporally grounded segmentation and causal explanation.
    \item \textit{\textbf{Large-scale Dataset:}} We present \textsc{DAR}, the first large-scale, viewer-centric emotion dataset based on Affective Events Theory, with 15{,}087 videos, event-aligned segments, and verified causal reasoning chains.
    \item \textit{\textbf{Construction Method:}} We develop a scalable data construction pipeline that combines coarse-to-fine event segmentation, incremental differential captioning, and multi-dimensional quality verification to produce dense and reliable annotations.
    \item \textit{\textbf{Strong Baseline:}} We propose \textsc{DAR-R1}, a two-stage tuning approach that leverages reinforcement learning to improve temporal boundary localization and reasoning quality, establishing a strong baseline for this benchmark.
\end{enumerate}

\section{Related Work}
\label{sec:related_work}

\subsection{Video Emotion Analysis}
Traditional Video Emotion Analysis~\cite{ye2025multi,song2024emotional} is often formulated as static emotion recognition on short clips or constrained scenarios~\cite{jiang2020dfew}, where a video segment is assigned a single categorical label~\cite{cao2014crema} or emotions are inferred from affective facial cues~\cite{kollias2018aff}.
While these benchmarks have driven progress in classification, they typically provide coarse temporal supervision and rarely evaluate whether a model can explain \emph{why} an emotion arises or \emph{when} it shifts.
With the emergence of MLLMs, recent datasets and benchmarks have pushed VEA toward richer, language-based evaluation: AffectGPT~\cite{lian2025affectgpt} promotes descriptive emotion understanding at scale, and MME-Emotion~\cite{zhang2025mme} benchmarks both recognition and emotion-related reasoning.
Despite these advances, fine-grained temporal localization and event-level causal attribution remain under-explored in standard VEA settings.

\subsection{viewer-Centric Affective Reasoning}
Beyond character-centric recognition, viewer-centric affect modeling studies emotions \emph{induced in the observer} by visual stimuli.
For images, ArtEmis~\cite{ArtEmisv1} collects large-scale emotion attributions with free-form explanations for artworks, ArtEmis v2.0~\cite{ArtEmisv2} reduces emotional bias via contrastive data collection, and Affection~\cite{Affection} scales affective explanations to real-world images.
For videos, VCE provides viewer-centric emotion annotations, highlighting the subjectivity of audience reactions, while StimuVAR~\cite{guo2025stimuvar} formulates video affective reasoning by predicting viewer emotions and generating rationales with stimuli-aware modeling.
However, these resources often operate at the image level or provide video-level affect without dense, event-aligned boundaries, making it difficult to benchmark fine-grained affect transitions. In contrast, \textsc{DAR} provides event-aligned, temporally grounded annotations with explicit causal supervision for dynamic affective reasoning.

\subsection{Multimodal Large Language Models}
MLLMs connect visual encoders with LLM backbones and are adapted via instruction tuning~\cite{xie2024emovit}, while video-oriented MLLMs extend the interface to temporal inputs~\cite{maaz2024video,su2023pandagpt,lin2024video}.
Post-training methods, including reinforcement learning~\cite{schulman2017proximal} and group-based optimization~\cite{guo2025deepseek}, have been explored to improve structured outputs and reasoning quality.
Importantly, this paper does not focus on designing new policy optimization algorithms; instead, we emphasize crafting task-specific reward signals for fine-grained emotion recognition and dynamic affective reasoning.

\begin{table}[t]
    \centering
    \caption{\textbf{Comparison of Video Emotion Datasets.} Descriptive datasets allow for more diverse labels, facilitating the modeling of fine-grained dynamic emotions.}
    \label{tab:dataset_comparison}
    \setlength{\tabcolsep}{4pt}  
    \resizebox{\textwidth}{!}{
    \begin{tabular}{l | c c c c c c c}  
    \toprule
    \textbf{Dataset} & \textbf{Video} & \textbf{Data} & \textbf{Classes} & \textbf{Description} & \textbf{Reason} & \textbf{Timestamp} & \textbf{Dynamic Emotion}\\
    \midrule
    DFEW~\cite{jiang2020dfew} & 16,372 & 16,372 & 7 & \ding{55} & \ding{55} & \ding{55}& \ding{55} \\
    CREMA-D~\cite{cao2014crema} & 7,442 & 7,442 & 6 & \ding{55} & \ding{55} & \ding{55} & \ding{55}\\
    FERV39k~\cite{wang2022ferv39k} & 38,935 & 38,935 & 7 & \ding{55} & \ding{55} & \ding{55} & \ding{55}\\
    MER2023~\cite{lian2023mer} & 5,030 & 5,030 & 6 & \ding{55} & \ding{55} & \ding{55} & \ding{55}\\
    VCE~\cite{mazeika2022would} & 61,046 & 61,046 & 27 & \ding{55} & \ding{55} & \ding{55} & \ding{55}\\
    MELD~\cite{poria2019meld} & 13,708 & 13,708 & 7 & \ding{55}  & \ding{55} & \textbf{\checkmark} & \ding{55}\\
    MERR~\cite{cheng2024emotion} & 28,618 & 28,618 & 113 & \checkmark & \ding{55} & \ding{55} & \ding{55}\\
    MER-Caption+~\cite{lian2025affectgpt} & 31,327 & 31,327 & 1,972 & \checkmark & \ding{55} & \ding{55} & \ding{55}\\
    \midrule
    \textbf{DAR (Ours)} & 15,087 & 36,908 & 27 & \textbf{\checkmark} & \textbf{\checkmark} & \textbf{\checkmark} & \textbf{\checkmark}\\
    \bottomrule
    \end{tabular}
    }
\end{table}

\section{The \textsc{DAR} Dataset}
\label{sec:dataset}

Table~\ref{tab:dataset_comparison} summarizes representative video emotion datasets, most of which frame emotion recognition as static classification over short clips, overlooking the temporal dynamics and causal nature of affective experience. To bridge this gap, we introduce \textsc{DAR}, a large-scale, viewer-centric dataset designed for dynamic affective reasoning. In this section, we first outline the theoretical foundation grounding our construction philosophy. We then detail our construction pipeline, quality assurance protocols, and dataset statistics.

\begin{figure}[t]
  \centering
  \includegraphics[width=0.99\linewidth]{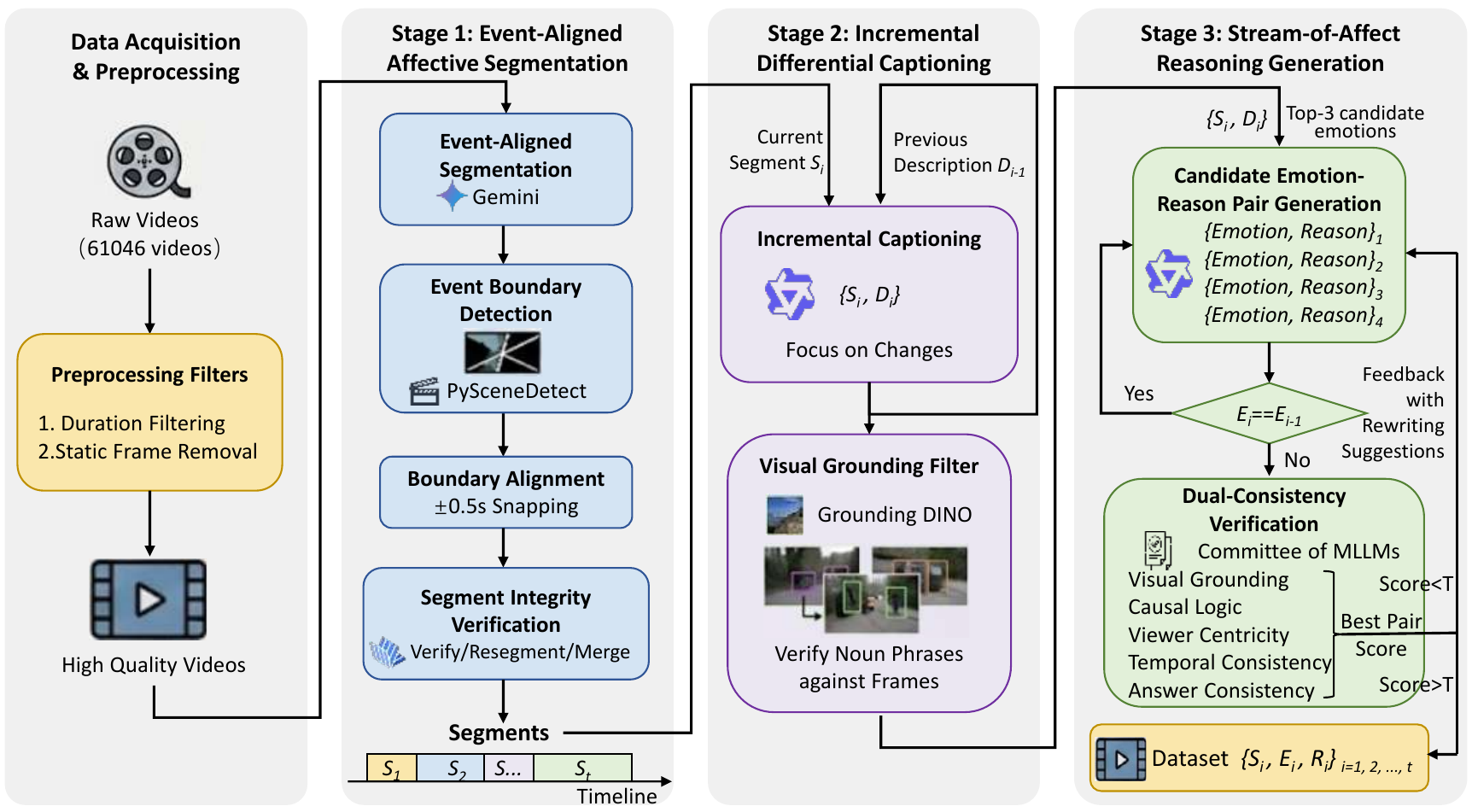}
  \caption{\textbf{Data Construction Pipeline of DAR.} The process consists of three stages: (1) Coarse-to-fine event segmentation combining semantic proposals with visual cuts; (2) Incremental differential captioning to reduce redundancy; and (3) Stream-of-affect reasoning generation grounded in AET. A dual-consistency verification protocol ensures annotation quality.}
  \label{fig:pipeline}
\end{figure}

\subsection{Theoretical Foundation: Affective Events Theory}
Our dataset construction is grounded in the \textit{Affective Events Theory} (AET) proposed by Weiss and Cropanzano~\cite{weiss1996affective}. AET posits that human emotions are not static states, but rather dynamic reactions triggered by consecutive, precipitating events—such as sudden visual stimuli or narrative twists. Furthermore, these affective reactions are dynamic and cumulative, meaning current emotions are influenced by the history of preceding events. Based on this theory, we construct our dataset by segmenting videos based on semantic and visual event boundaries rather than arbitrary time intervals. This ensures that each segment represents a coherent \textit{stimulus unit} that triggers a specific \textit{affective reaction}, providing a solid psychological basis for dynamic reasoning. 

\subsection{Data Acquisition and Preprocessing}
We source our raw videos from the Video Cognitive Empathy (VCE) dataset~\cite{mazeika2022would}, which provides a rich taxonomy of 27 fine-grained emotion categories suitable for capturing nuanced viewer reactions. 

\noindent\textbf{Duration Filtering.} We discard videos shorter than 1 second, as they lack sufficient temporal context to exhibit dynamic emotional changes.

\noindent\textbf{Static Frame Removal.} We compute the perceptual hash and optical flow between frames. Videos exhibiting static imagery, such as slideshows or single images, are removed as they do not possess the temporal motion cues essential for analyzing event-driven emotional shifts.

\subsection{Construction Pipeline}
Our construction pipeline, as illustrated in Figure~\ref{fig:pipeline}, transforms raw videos into temporally grounded, causal affective reasoning chains through a multi-stage process.

\noindent\textbf{Stage 1: Event-Aligned Affective Segmentation.}
To implement the AET-driven segmentation, we employ a coarse-to-fine strategy combining semantic understanding with visual boundary detection. First, we prompt Gemini-2.5-Pro~\cite{comanici2025gemini} to propose high-level semantic event boundaries, dividing the video according to narrative shifts or affective turning points. Since LLM-predicted timestamps often suffer from drift, we utilize PySceneDetect~\cite{castellano2025pyscenedetect} to detect precise hard cuts and fade transitions. We apply boundary snapping: semantic boundaries within a 0.5\,s window of a detected visual cut are shifted to the corresponding cut frame. Finally, for segments that do not align with hard cuts, we employ InternVL3.5~\cite{wang2025internvl3} to verify segment integrity, recursively re-segmenting or merging those that fail to represent a complete event.

\noindent\textbf{Stage 2: Incremental Differential Captioning.}
Conventional video captioning often yields redundant information when applied to streaming segments. Let $S_i$ denote the $i$-th segment in a video, and let $D_i$ denote its visual description. For the first segment, $D_0$ is set to an empty context. We introduce an \textit{Incremental Differential Captioning} strategy in which Qwen3-VL~\cite{bai2025qwen3} is conditioned on the visual tokens of $S_i$ and the preceding description $D_{i-1}$. The model is explicitly instructed to focus on the changes in visual content, actions, and atmosphere. To ensure visual faithfulness, we use Grounding DINO~\cite{liu2024grounding} as a visual grounding filter to verify whether salient entities mentioned in each description can be grounded in the associated video frames.

\noindent\textbf{Stage 3: Stream-of-Affect Reasoning Generation.}
This stage generates the core reasoning chain linking stimuli to emotions. We adopt a ``Bottom-up'' reasoning logic: \textit{Visual Evidence $\rightarrow$ Contextual Appraisal $\rightarrow$ Emotion Trigger}. Let $E_i$ denote the viewer-centric emotion label associated with segment $S_i$. For the first segment, $E_0$ is set to an empty context. We construct a structured prompt containing the historical context $\{S_{i-1}, D_{i-1}, E_{i-1}\}$, the current stimulus $\{S_i, D_i\}$, and the Top-3 candidate emotions drawn from VCE, and provide it to Qwen3-VL. The model generates four candidate \texttt{emotion}--\texttt{reason} pairs. Following AET principles, if the predicted emotion for segment $S_i$ is identical to that of $S_{i-1}$, we merge the two segments into a coherent emotional phase and regenerate the rationale to reflect the sustained affective state.

\subsection{Quality Assurance: Dual-Consistency Verification}
To ensure the quality of our model-generated data, we implement a Dual-Consistency Verification Protocol. We employ InternVL3.5 and Qwen3-omni~\cite{xu2025qwen3} as two independent MLLM judges. Instead of binary scoring, the generated \texttt{emotion} and \texttt{reason} pairs are evaluated across five distinct dimensions:

\begin{enumerate}
    \item \textbf{Visual Grounding.} This metric verifies whether the visual cues cited in the reasoning, such as specific objects or actions, are physically present in the corresponding video frames.
    \item \textbf{Causal Logic.} This dimension assesses whether the described event provides a sufficient and logical premise to trigger the predicted emotion.
    \item \textbf{Viewer Centricity.} This criterion ensures the reasoning focuses on the viewer's affective reaction rather than merely describing the character's facial expressions or internal state.
    \item \textbf{Temporal Consistency.} This metric checks if the reasoning coheres with the historical context of previous segments, which ensures a continuous narrative flow.
    \item \textbf{Answer Consistency.} This verifies whether the final emotion label logically follows from the arguments presented in the reasoning text.
\end{enumerate}

The committee assigns a comprehensive score based on these dimensions. Low-scoring samples are not simply discarded; they are fed back into the generation pipeline with specific feedback for rewriting. Only samples that pass a strict quality threshold after verification are included in the final dataset.

\begin{figure}[t]
  \centering
  \captionsetup[subfigure]{skip=2.5pt}

  \begin{subfigure}[b]{0.48\linewidth}
    \centering
    \includegraphics[
      width=\linewidth,
      height=0.17\textheight,
      keepaspectratio
    ]{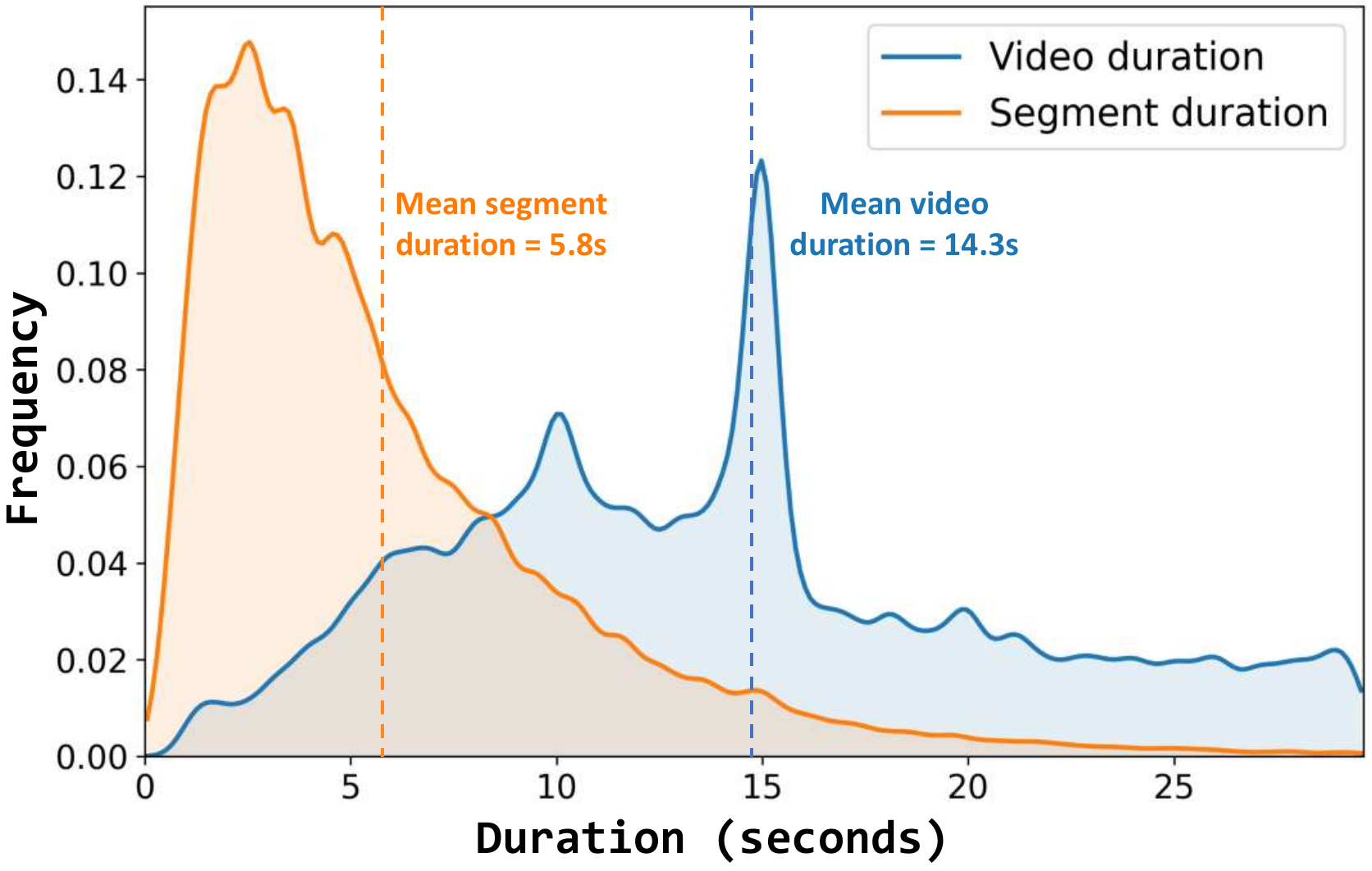}
    \caption{}
    \label{fig:sub-a}
  \end{subfigure}\hfill
  \begin{subfigure}[b]{0.48\linewidth}
    \centering
    \includegraphics[
      width=\linewidth,
      height=0.17\textheight,
      keepaspectratio
    ]{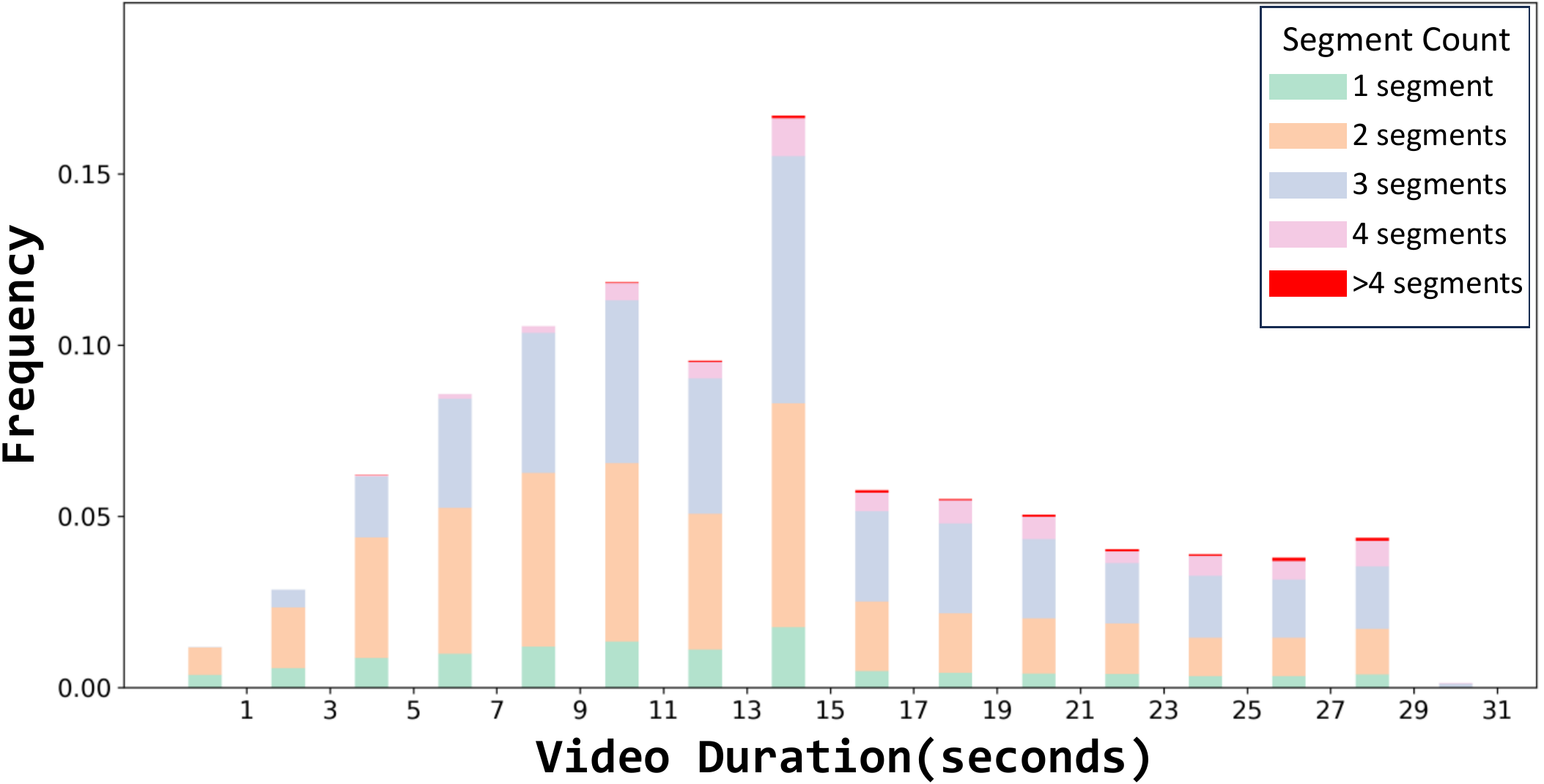}
    \caption{}
    \label{fig:sub-b}
  \end{subfigure}

  \begin{subfigure}[b]{0.48\linewidth}
    \centering
    \includegraphics[
      width=\linewidth,
      height=0.2\textheight,
      keepaspectratio
    ]{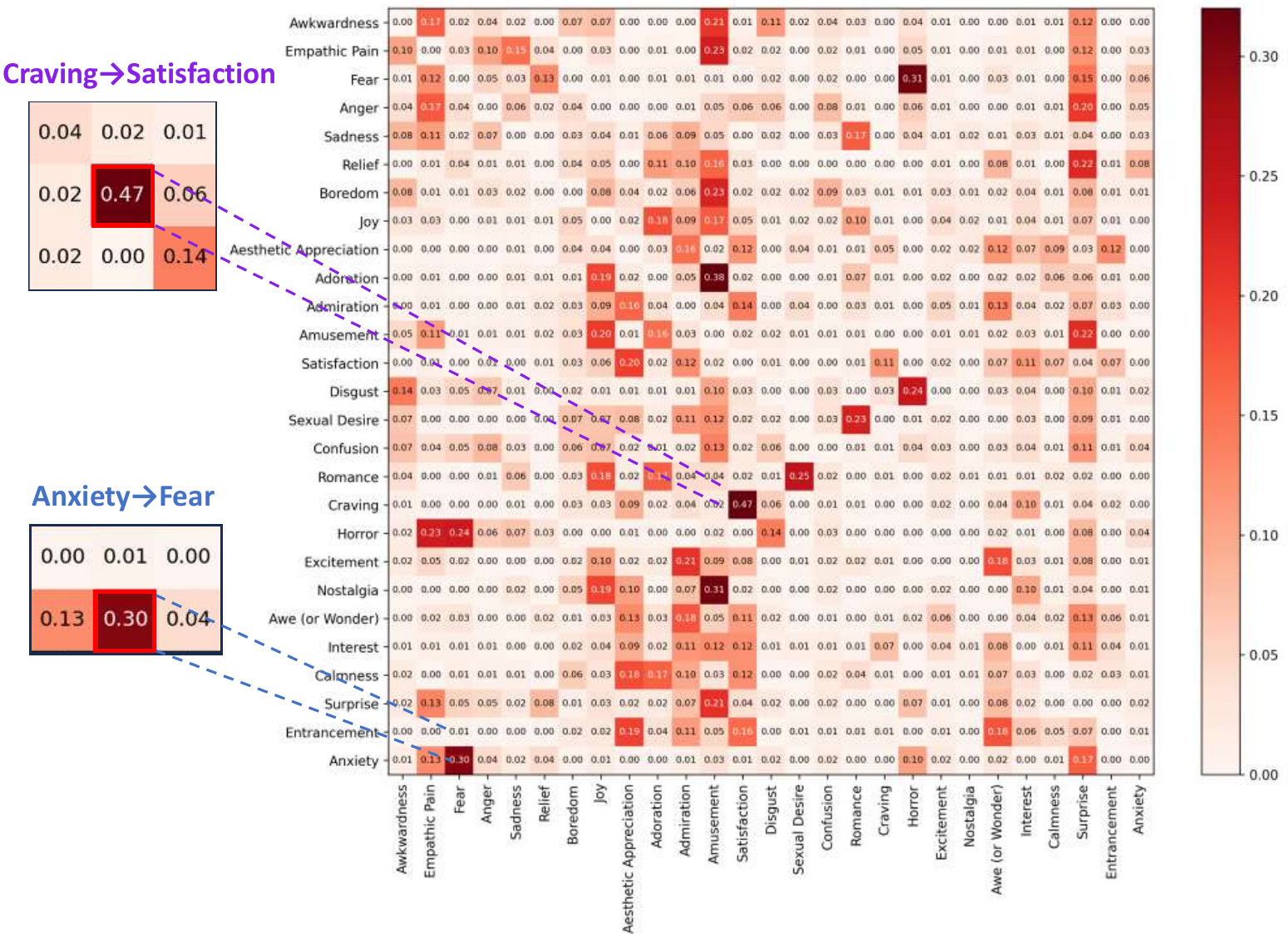}
    \caption{}
    \label{fig:sub-c}
  \end{subfigure}\hfill
  \begin{subfigure}[b]{0.48\linewidth}
    \centering
    \includegraphics[
      width=\linewidth,
      height=0.2\textheight,
      keepaspectratio
    ]{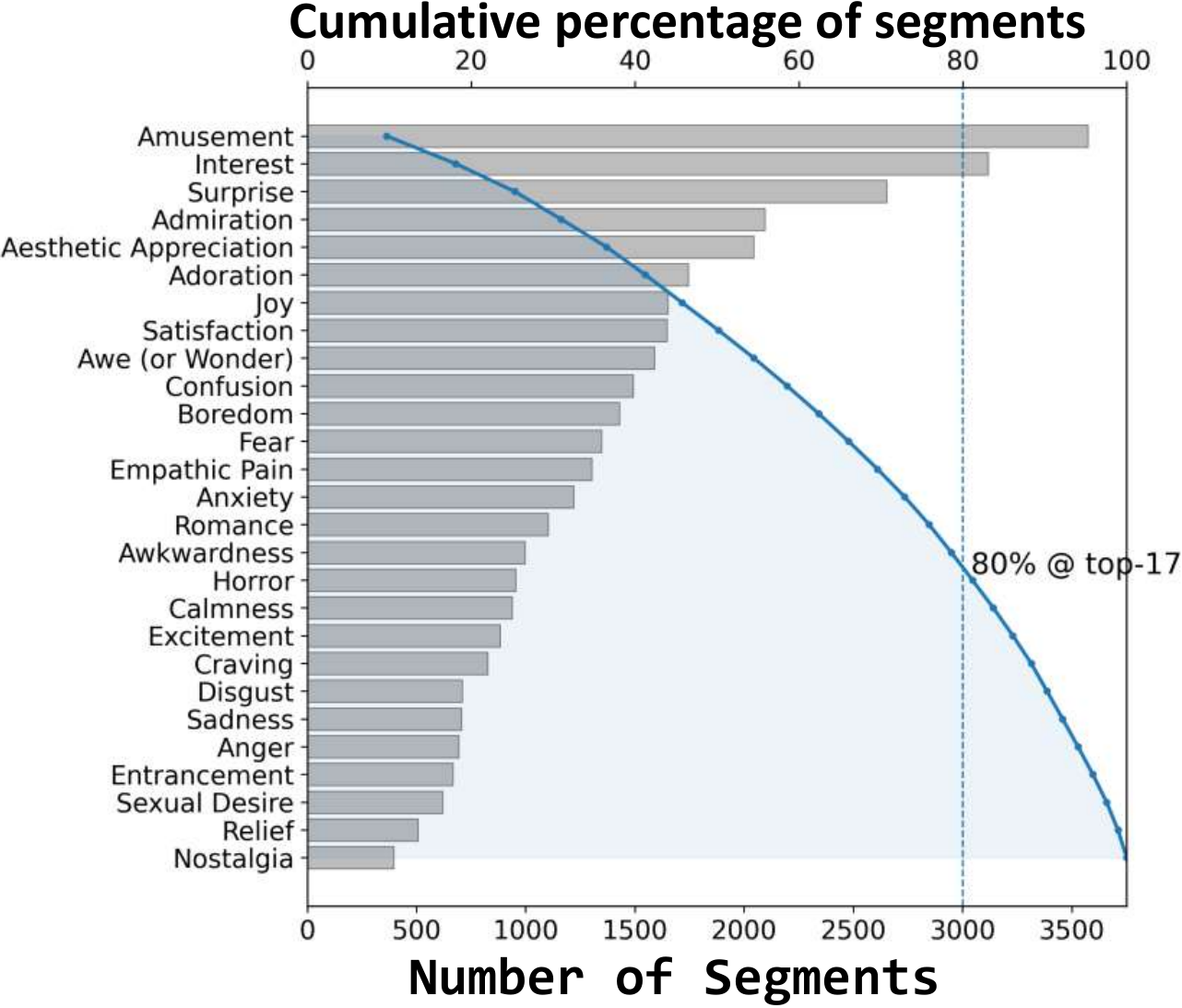}
    \caption{}
    \label{fig:sub-d}
  \end{subfigure}

  \begin{subfigure}[b]{0.48\linewidth}
    \centering
    \includegraphics[
      width=\linewidth,
      height=0.17\textheight,
      keepaspectratio
    ]{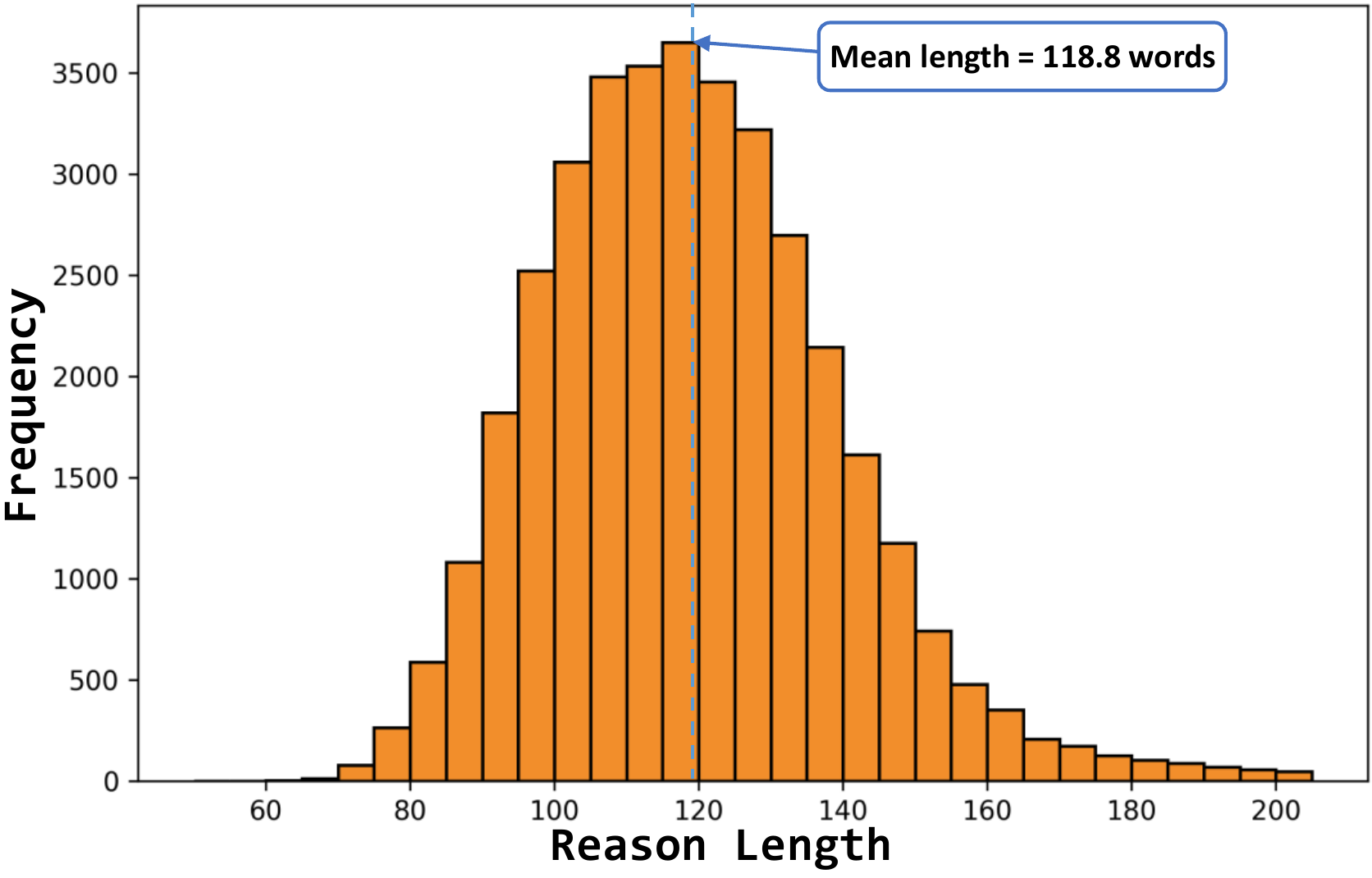}
    \caption{}
    \label{fig:sub-e}
  \end{subfigure}\hfill
  \begin{subfigure}[b]{0.48\linewidth}
    \centering
    \includegraphics[
      width=\linewidth,
      height=0.17\textheight,
      keepaspectratio
    ]{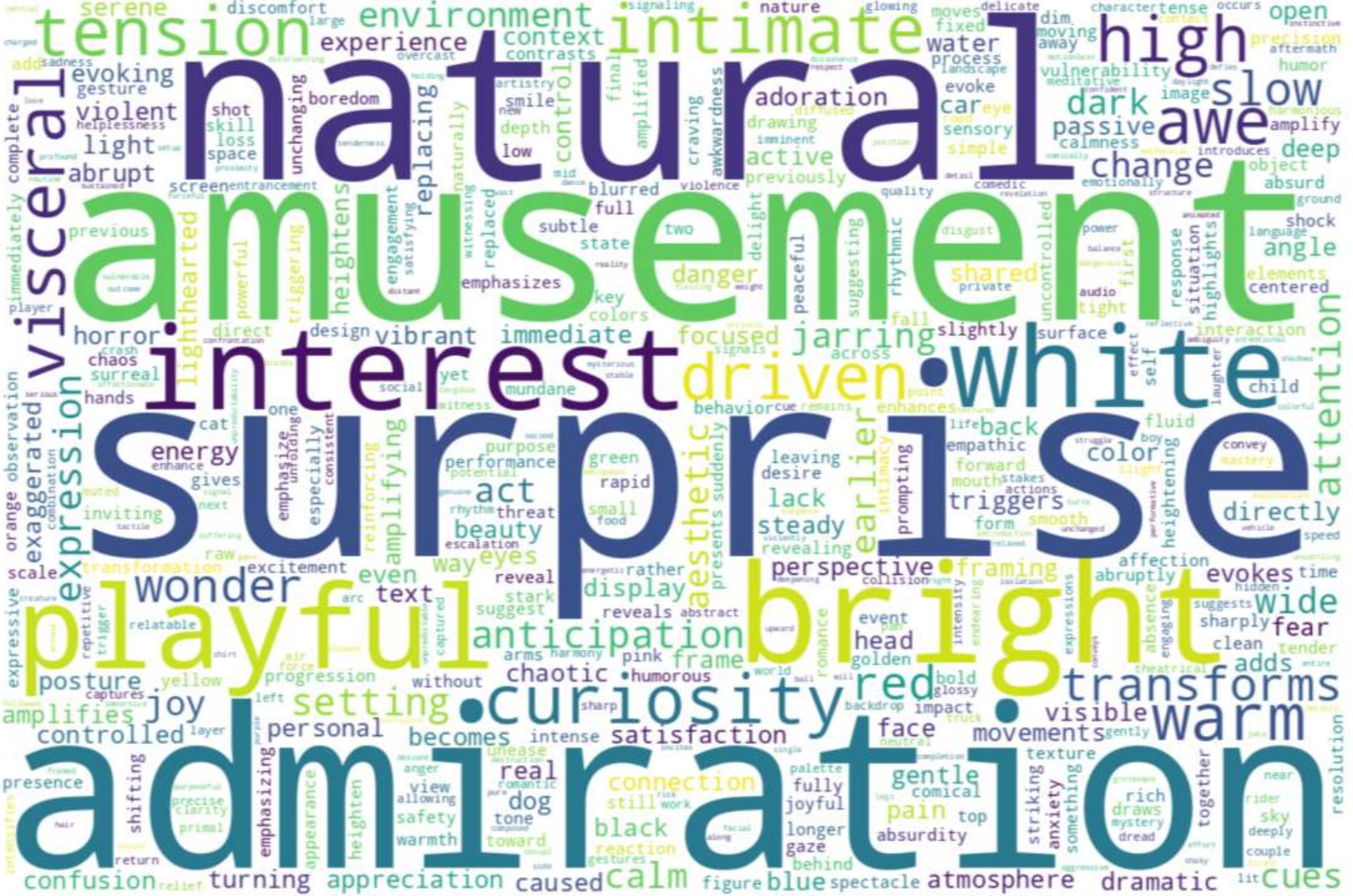}
    \caption{}
    \label{fig:sub-f}
  \end{subfigure}

  \caption{\textbf{Statistical Analysis of the \textsc{DAR} Dataset.}
  (a) Video and segment duration distributions.
  (b) Segment counts by video duration.
  (c) Emotion transition matrix.
  (d) Distribution of 27 emotion categories.
  (e) Reasoning-text lengths.
  (f) High-frequency words in reasoning annotations.}
  \label{fig:stats}
\end{figure}

\subsection{Dataset Statistics}
We provide a comprehensive statistical analysis of DAR to highlight its scale, diversity, and complexity. The dataset contains 15,087 videos and 36,908 affective segments, with 13,646 videos in the training set and 1,441 videos in the test set.

\noindent\textbf{Temporal Granularity and Dynamics.}
Unlike previous datasets that treat videos as static, \textsc{DAR} emphasizes fine-grained temporal dynamics. As shown in Figure~\ref{fig:stats}(a), the average video duration is 14.3s, while the average segment duration is 5.8s, indicating that a typical video contains multiple distinct emotional phases. This is further corroborated by Figure~\ref{fig:stats}(b), which illustrates the relationship between video length and segment count, reinforcing the idea that emotions can evolve quickly within brief video durations. 
Furthermore, the emotion transition matrix in Figure~\ref{fig:stats}(c) reveals structured and non-random affect dynamics, with the most prominent transition being \textit{Craving} $\rightarrow$ \textit{Satisfaction} (0.47) and another frequent shift from \textit{Anxiety} $\rightarrow$ \textit{Fear} (0.30), indicating that our annotations capture coherent, event-driven emotional progressions rather than arbitrary fluctuations.

\noindent\textbf{Emotion Distribution.}
Figure~\ref{fig:stats}(d) presents the distribution of the 27 emotion categories. While the distribution follows a natural long-tail pattern typically observed in real-world data, the top-17 emotions account for approximately 80\% of the occurrences. This indicates a balanced emotional coverage within the dataset, where emotions such as Amusement, Interest, and Surprise dominate, while complex emotions like Admiration and Aesthetic Appreciation are well-represented. This demonstrates that our dataset not only captures basic Ekman emotions but also includes more nuanced emotional and cognitive appraisals.

\noindent\textbf{Reasoning Complexity.}
The average reasoning text length, as shown in Figure~\ref{fig:stats}(e), is 118.8 words, which is significantly longer than standard video captions. This suggests that our annotations provide rich contextual information rather than merely summarizing the content. 
The word cloud in Figure~\ref{fig:stats}(f) shows the high-frequency affect-related terms in the reasoning annotations after filtering generic nouns and connective words. The remaining terms cover emotion words, appraisal-related expressions, and visual descriptors.

\section{Methodology}
\label{sec:method}

To address the challenges of dynamic affective reasoning—specifically, the need for precise temporal localization, accurate emotion classification, and grounded causal reasoning—we propose DAR-R1. Our framework adapts an MLLM to this task via a two-stage training paradigm: cold-start supervised fine-tuning (SFT) for structural adaptation and reinforcement learning (RL) for reasoning refinement.

\subsection{Cold Start Training}
\label{subsec:sft}
The primary objective of this stage is to adapt the generalist MLLM to the complex output schema required for dynamic reasoning and to initialize a policy prior for the subsequent RL stage.

Given visual inputs $\mathcal{V}$ and task instructions $\mathcal{I}_{\mathrm{task}}$, the model is trained to generate the ground-truth target sequence $\mathcal{S}_{gt}=(S_1^*,\ldots,S_N^*)$, where each segment $S_i^{*}=(t_{i,\mathrm{start}}^{*}, t_{i,\mathrm{end}}^{*}, e_i^{*}, r_i^{*})$ comprises the start and end timestamps, emotion label, and causal rationale. We optimize the parameters $\theta$ by minimizing the standard negative log-likelihood over the ground-truth tokens.

\subsection{Reinforcement Learning via GRPO}
\label{subsec:rl}
While SFT ensures format compliance, it often struggles to capture precise event boundaries or generate deeply grounded reasoning. To this end, we employ Group Relative Policy Optimization~\cite{guo2025deepseek} (GRPO). Unlike Proximal Policy Optimization~\cite{schulman2017proximal} (PPO) which requires a separate critic model, GRPO estimates the baseline from a group of outputs generated by the current policy itself, reducing computational overhead.
For each input query $q$, we sample a group of $G$ outputs
$\{o_i\}_{i=1}^{G}$ from the behavior policy
$\pi_{\theta_{\mathrm{old}}}$. The objective function aims to maximize the advantage of high-reward outputs while constraining the policy shift via KL divergence:
\begin{equation}
\mathcal{J}_{\mathrm{GRPO}}(\theta)=
\mathbb{E}_{q\sim\mathcal{D},\,\{o_i\}_{i=1}^G\sim\pi_{\theta_{\text{old}}}(\cdot|q)}
\left[\frac{1}{G}\sum_{i=1}^G \mathcal{G}_i(\theta)\right],
\end{equation}
\begin{equation}
\mathcal{G}_i(\theta)=
\frac{1}{|o_i|}\sum_{t=1}^{|o_i|}
\min\!\Big(
\rho_{i,t}(\theta)\,\hat A_i,\;
\mathrm{clip}(\rho_{i,t}(\theta),1-\epsilon,1+\epsilon)\,\hat A_i
\Big)
-\beta\,\mathbb{D}_{\mathrm{KL}}\!\left(\pi_\theta\,\|\,\pi_{\mathrm{ref}}\right),
 \end{equation}
\begin{equation}
\rho_{i,t}(\theta)=
\frac{\pi_\theta(o_{i,t}\mid q,o_{i,<t})}{\pi_{\theta_{\text{old}}}(o_{i,t}\mid q,o_{i,<t})},
\quad
\hat{A}_i = \frac{R_i - \text{mean}(R_{group})}{\text{std}(R_{group})}.
\end{equation}

We design a multi-dimensional reward function to guide the model across structure, localization, accuracy, and reasoning. 

\noindent\textbf{Structural Constraints Reward ($R_{struct}$).}
This component ensures the output adheres to the strict JSON schema, maintains temporal validity, and prevents runaway generation. It aggregates three sub-rewards: format compliance ($r_{fmt}$), minimum segment duration ($r_{dur}$), and total length penalty ($r_{len}$). Let $o$ denote the generated output text, and $\mathcal{S}_{pred} = (\hat S_1,\ldots,\hat S_N)$ be the set of parsed segments, where each segment $\hat S_i$ contains a start time $\hat t_{i,\mathrm{start}}$ and end time $\hat t_{i,\mathrm{end}}$. The structural reward is defined as:
\begin{equation}
    R_{struct} = \lambda_1 \cdot \mathbb{I}(o \in \mathcal{J}) + \lambda_2 \cdot \prod_{\hat S_i \in \mathcal{S}_{pred}} \mathbb{I}(\hat t_{i,\mathrm{end}} - \hat t_{i,\mathrm{start}} \ge \tau_{min}) + \lambda_3 \cdot \psi_{len}(|o|),
\end{equation}
where $\mathbb{I}(\cdot)$ is the indicator function, and $\mathcal{J}$ represents the set of valid JSON strings that conform to the required schema (e.g., valid keys, continuous timestamps). We set $\lambda_1=0.5$, $\lambda_2=0.2$, and $\lambda_3=0.3$. The second term penalizes unrealistically short segments ($\tau_{min}=0.4s$) to avoid fragmentation. The third term $\psi_{len}(|o|)$ applies a soft penalty on the total token count $|o|$ to discourage excessively long or runaway generations, defined as $\psi_{len}(n) = \exp(-\max(0, n - L_{soft})/\tau_{len})$ if $n < L_{hard}$, and 0 otherwise. We use $L_{soft}=1200$, $L_{hard}=2400$, and $\tau_{len}=400$.

\noindent\textbf{Segment Count Reward ($R_{count}$).}
To prevent over-segmentation or under-segmentation, we encourage the model to generate a number of segments consistent with the ground truth. Let $N_{pred} = |\mathcal{S}_{pred}|$ and $N_{gt} = |\mathcal{S}_{gt}|$, respectively. The reward is modeled as an exponential decay function based on the count difference:
\begin{equation}
    R_{count} = \exp\left(-\frac{|N_{pred} - N_{gt}|}{\tau_{cnt}}\right),
\end{equation}
where $\tau_{\rm cnt}=1.5$ controls the sensitivity of the penalty. This formulation provides a dense reward signal, incentivizing predictions that closely approximate the ground-truth event density.

\noindent\textbf{Temporal Segmentation Reward ($R_{seg}$).}
To improve boundary precision, we employ an Intersection over Union (IoU) matching mechanism combined with a boundary distance decay. For each predicted segment $\hat S_i \in \mathcal{S}_{pred}$, we identify the best-matching ground-truth segment $S_i^* \in \mathcal{S}_{gt}$ that maximizes the temporal IoU. The segment-level reward is then computed as:
\begin{equation}
    r_{seg}(\hat S_i) = \alpha \cdot \text{IoU}(\hat S_i, S_i^*) + (1-\alpha) \cdot \exp\left(-\frac{|\hat t_{i,\mathrm{start}} - t_{i,\mathrm{start}}^*| + |\hat t_{i,\mathrm{end}} - t_{i,\mathrm{end}}^*|}{\tau_{temp}}\right).
\end{equation}
The final $R_{seg}$ is the average of $r_{seg}$ over all predicted segments. The first term incentivizes overlap, while the second term, governed by the boundary decay temperature $\tau_{\rm temp}=0.5$, encourages the start and end timestamps to snap precisely to visual event boundaries. We use $\alpha=0.5$ to balance temporal overlap and boundary-distance accuracy.

\noindent\textbf{Emotion Accuracy Reward ($R_{emo}$).}
This reward evaluates the correctness of the predicted emotion labels. We align predicted segments to ground truth based on maximum temporal overlap. Let $\hat e_i$ and $e_i^*$ be the emotion labels for the matched segments $\hat S_i$ and $S_i^*$. The reward is binary, conditioned on both label matching and sufficient temporal overlap:
\begin{equation}
    R_{emo} = \frac{1}{N_{pred}} \sum_{\hat S_i \in \mathcal S_{\mathrm{pred}}} \mathbb{I}(\hat e_i = e_i^*) \cdot \mathbb{I}(\text{IoU}(\hat S_i, S_i^*) > 0.5).
\end{equation}
This ensures that the model is rewarded only when it correctly identifies the emotion for the correct temporal event.

\noindent\textbf{Reasoning Quality Reward ($R_{reason}$).}
To ensure the generated rationale $r_i$ is informative and non-repetitive, we define $R_{reason}$ based on length and uniqueness constraints. Let $W(r_i)$ be the word count of the reasoning text for segment $S_i$. We define an exponential length score $\phi(l) = \exp(-|l - \mu_{len}| / \sigma_{len})$. The uniqueness is measured using the Jaccard similarity between adjacent reasoning texts $r_i$ and $r_{i-1}$. The reward is computed as:
\begin{equation}
    R_{reason} = \frac{1}{N_{pred}} \sum_{i=1}^{N_{pred}} \phi(W(\hat r_i)) \cdot (1 - \mathbb{I}(\text{Jaccard}(\hat r_i, \hat r_{i-1}) > \tau_{dup})),
\end{equation}
where $\mu_{len}=120$ encourages detailed explanations, $\sigma_{len}=40$ controls the length-score decay, and $\tau_{dup}=0.75$ serves as a threshold to penalize repetitive hallucination loops across consecutive segments.

The total reward $R_{total}$ is a weighted sum of five components:
\begin{equation}
    R_{total} = w_{1} R_{struct} + w_{2} R_{count} + w_{3} R_{seg} + w_{4} R_{emo} + w_{5} R_{reason},
\end{equation}
where $w_k$ represents the weight for each reward component.

\section{Experiments}
\label{sec:experiments}

In this section, we validate the effectiveness of our proposed \textsc{DAR-R1} framework on the \textsc{DAR} dataset. We first describe the experimental setup, including implementation details and evaluation metrics. We then present a comprehensive comparison against state-of-the-art Multimodal Large Language Models. Finally, we conduct ablation studies to analyze the impact of our two-stage training paradigm and provide a qualitative analysis.

\subsection{Experimental Setup}

\textbf{Implementation Details.}
We utilize Qwen2.5-VL-3B as our backbone model, keeping the vision encoder frozen while fine-tuning the LLM and aligner throughout both stages. We adopt a two-phase training scheme on 4$\times$H100 GPUs: in the cold-start stage, we run supervised fine-tuning for 0.5 epochs using AdamW with a learning rate of $1\times10^{-5}$; in the reinforcement learning stage, we perform GRPO training for 1 epoch with a learning rate of $2\times10^{-6}$. The overall reward is a weighted sum of structural constraints, segment count reward, temporal segmentation, emotion accuracy, and reasoning quality with weights 0.10, 0.25, 0.25, 0.25, and 0.15, respectively.

\noindent\textbf{Emotion Taxonomy.}
To capture fine-grained viewer reactions, our dataset utilizes the VCE taxonomy consisting of 27 emotion categories, including \textit{Admiration, Amusement, Anger, Anxiety, Awe, Boredom, Calmness, Confusion, Craving, Disgust, Empathic Pain, Entrancement, Excitement, Fear, Horror, Interest, Joy, Nostalgia, Relief, Romance, Sadness, Satisfaction, Sexual Desire, Surprise, Awkwardness, Adoration, Aesthetic Appreciation}.

\noindent\textbf{Evaluation Metrics.}
Evaluating dynamic affective reasoning requires assessing both temporal localization and semantic correctness. We adopt a holistic suite of metrics:
\begin{itemize}
    \item \textbf{Segment Count Accuracy:} We evaluate whether the number of predicted segments correctly reflects the emotional transitions in the video. This metric ensures that the predicted number of emotional shifts aligns with the ground truth, addressing the accuracy of emotional boundary detection.
    \item \textbf{Segmentation Accuracy:} We calculate the Mean Temporal Intersection over Union (mIoU) between the predicted segments and ground-truth segments to evaluate temporal localization accuracy.
    \item \textbf{Emotion Accuracy:} A prediction is considered correct only if the predicted emotion category matches the ground truth and the segment has a temporal IoU $\ge 0.5$ with the corresponding ground truth segment.
    \item \textbf{Reasoning Quality:} To evaluate rationale quality, we employ GPT-4o~\cite{hurst2024gpt} as a judge. The judge rates each generated rationale on a 0--5 scale, focusing on visual grounding, causal logic, viewer centricity, temporal consistency, and answer consistency.
\end{itemize}

\subsection{Main Results}

\begin{table}[t]
    \centering
    \setlength{\tabcolsep}{6pt}
    \captionsetup{font=footnotesize}
    \caption{\textbf{Quantitative Comparison on the DAR Test Set.} We compare our method with leading proprietary and open-source MLLMs. SC-Acc (\%) denotes segment count accuracy, mIoU (\%) denotes temporal segmentation overlap, Emo-Acc (\%) denotes emotion prediction accuracy, and GPT-Score (0--5) denotes reasoning quality, as evaluated by GPT across five dimensions: visual grounding (VG), causal logic (CL), viewer centricity (VC), temporal consistency (TC), and answer consistency (AC), along with the average score (Avg). ``DAR-SFT'' is after SFT and ``DAR-R1'' is the final model. Best results are in \textbf{bold}.}    
    \label{tab:main_results}
    \resizebox{\textwidth}{!}{
    \begin{tabular}{l c c c c c c c c c}
    \toprule
    \multirow{2}{*}{\textbf{Models}} &
    \multirow{2}{*}{\textbf{SC-Acc}} &
    \multirow{2}{*}{\textbf{mIoU}} &
    \multirow{2}{*}{\textbf{Emo-Acc}} &
    \multicolumn{6}{c}{\textbf{GPT-Score}} \\
    \cmidrule(lr){5-10}
     &  &  &  & \textbf{VG} & \textbf{CL} & \textbf{VC} & \textbf{TC} & \textbf{AC} & \textbf{Avg}\\
    \midrule
        \multicolumn{10}{c}{Emotion Focused Multimodal Large Language Models} \\
        \midrule
    Videmo~\cite{zhang2025videmo} & 7.9 & 9.2 & 4.3 & 0.7 & 0.6 & 0.5 & 0.4 & 0.3 & 0.5 \\
    EmotionLlama~\cite{cheng2024emotion} & 21.3 & 10.8 & 2.1 & 1.0 & 0.9 & 0.8 & 0.7 & 0.6 & 0.8 \\
    AffectGPT~\cite{lian2025affectgpt} & 24.3 & 11.9 & 5.8 & 1.0 & 1.4 & 1.3 & 0.7 & 1.1 & 1.1 \\
    \midrule
    \multicolumn{10}{c}{General Multimodal Large Language Models} \\
    \midrule
        Video-ChatGPT~\cite{maaz2024video} & 18.4 & 20.9 & 6.4 & 2.4 & 1.2 & 0.8 & 1.6 & 1.3 & 1.5 \\
        PandaGPT~\cite{su2023pandagpt}     & 21.3 & 32.7 & 7.1 & 1.9 & 2.4 & 1.2 & 2.2 & 1.5 & 1.8 \\
        Video-LLaVA~\cite{lin2024video}  & 14.6 & 32.6 & 7.0 & 2.1 & 1.9 & 1.0 & 1.8 & 1.7 & 1.7 \\
        InternVL-3.5-2B~\cite{wang2025internvl3} & 21.2 & 33.6 & 8.2 & 1.8 & 2.0 & 2.3 & 2.1 & 1.0 & 1.8 \\
        InternVL-3.5-8B~\cite{wang2025internvl3} & 26.0 & 39.2 & 11.5 & 2.2 & 2.5 & 2.5 & 2.3 & 1.6 & 2.2 \\
        Qwen2.5-VL-3B~\cite{bai2025qwen25vltechnicalreport} & 25.4 & 41.7 & 15.0 & 2.4 & 2.3 & 2.2 & 2.5 & 2.0 & 2.3 \\
        Qwen2.5-VL-7B~\cite{bai2025qwen25vltechnicalreport} & 31.9 & 45.2 & 14.9 & 2.8 & 2.7 & 2.6 & 2.6 & 2.4 & 2.6 \\
        Qwen3-VL-4B~\cite{bai2025qwen3}  & 29.2 & 48.2 & 17.9 & 2.8 & 2.9 & 2.6 & 2.8 & 2.4 & 2.7 \\
        Qwen3-VL-8B~\cite{bai2025qwen3}  & 26.2 & 43.1 & 17.0 & 2.6 & 3.0 & 2.4 & 2.6 & 2.2 & 2.6 \\
        \midrule
        \textbf{DAR-SFT} & 37.7 & 47.6 & 25.8 & 3.1 & 3.0 & 2.9 & 2.8 & 3.0 & 3.0 \\
        \textbf{DAR-R1} & \textbf{41.5} & \textbf{52.3} & \textbf{28.6} & \textbf{3.1} & \textbf{3.8} & \textbf{3.2} & \textbf{3.2} & \textbf{3.1} & \textbf{3.3} \\
        \bottomrule
    \end{tabular}
    }
\end{table}

Table~\ref{tab:main_results} compares our models with representative state-of-the-art MLLMs on the \textsc{DAR} test set. The results highlight both the difficulty of Dynamic Affective Reasoning and the effectiveness of our two-stage training paradigm.

\noindent\textbf{Limitations of Existing Models.}
Existing models struggle significantly with this benchmark, reflecting the combined challenges of strict temporal localization and causally grounded, viewer-centric explanation. Emotion-focused MLLMs like Videmo and AffectGPT perform poorly due to their static training paradigms, lacking the granularity for temporal localization. Large-scale generalist models such as Qwen3-VL-8B demonstrate stronger inherent capabilities but still fall short in temporal precision and reasoning coherence without domain-specific adaptation.

Incorporating the DAR dataset via cold-start training yields a substantial performance improvement. Compared to the base model Qwen-2.5-VL-3B, DAR-SFT achieves gains across all evaluation metrics, with particularly large gains in segment-count accuracy and emotion accuracy by 12.3\% and 10.8\%, respectively. This confirms that our dataset effectively bridges the gap between general multimodal understanding and dynamic affective alignment. Building upon this, the RL stage drives further gains. DAR-R1 consistently outperforms \textsc{DAR-SFT} across all metrics, achieving 52.3\% mIoU and a causal-logic score of 3.8, and establishes a new state of the art on our benchmark. This trajectory indicates that while SFT establishes structural compliance, the specific reward modeling in RL is crucial for refining boundary precision and enforcing logical consistency in affective reasoning.

\begin{wraptable}{r}{0.58\textwidth}
\vspace{-20pt}
  \centering
    \caption{\textbf{Human Evaluation of Reasoning Quality.} Average expert ratings across the five predefined dimensions.}  \label{tab:model_scores}
    \begin{tabular}{lcccccc}
    \toprule
    \textbf{Models} & \textbf{VG} & \textbf{CL} & \textbf{VC} & \textbf{TC} & \textbf{AC} & \textbf{Avg} \\
    \midrule
    AffectGPT      & 0.6 & 1.8 & 0.2 & 0.1 & 1.0 & 0.7 \\
    Qwen2.5-VL-3B  & 2.5 & 2.3 & 2.5 & 2.4 & 2.8 & 2.5 \\
    Qwen3-VL-4B    & 3.1 & 3.5 & 2.7 & 2.8 & 3.0 & 3.0 \\
    DAR-SFT        & 3.4 & 3.5 & 4.0 & 3.1 & 3.8 & 3.6 \\
    DAR-R1         & \textbf{4.2} & \textbf{4.5} & \textbf{4.2} & \textbf{3.8} & \textbf{4.4} & \textbf{4.2} \\
    \bottomrule
  \end{tabular}
  \vspace{0pt}
\end{wraptable}
\noindent\textbf{Human Evaluation.} To complement automatic metrics and validate alignment with human perception, we recruited five experts to evaluate 100 randomly sampled videos. As summarized in Table~\ref{tab:model_scores}, existing emotion-focused models like AffectGPT struggle with the multi-task complexity. Generalist MLLMs exhibit better foundational reasoning but lack domain specificity. In contrast, our proposed methods demonstrate superior performance. The high scores in Causal Logic and Viewer Centricity particularly verify that our RL paradigm effectively aligns the model with the nuanced, viewer-centric logic required for high-quality affective reasoning. Importantly, we observe strong agreement between GPT-based judging and expert preferences, suggesting that our automatic GPT evaluation provides a reliable proxy for human assessment on this benchmark.

\begin{figure}[t]
    \centering
    \includegraphics[width=\linewidth]{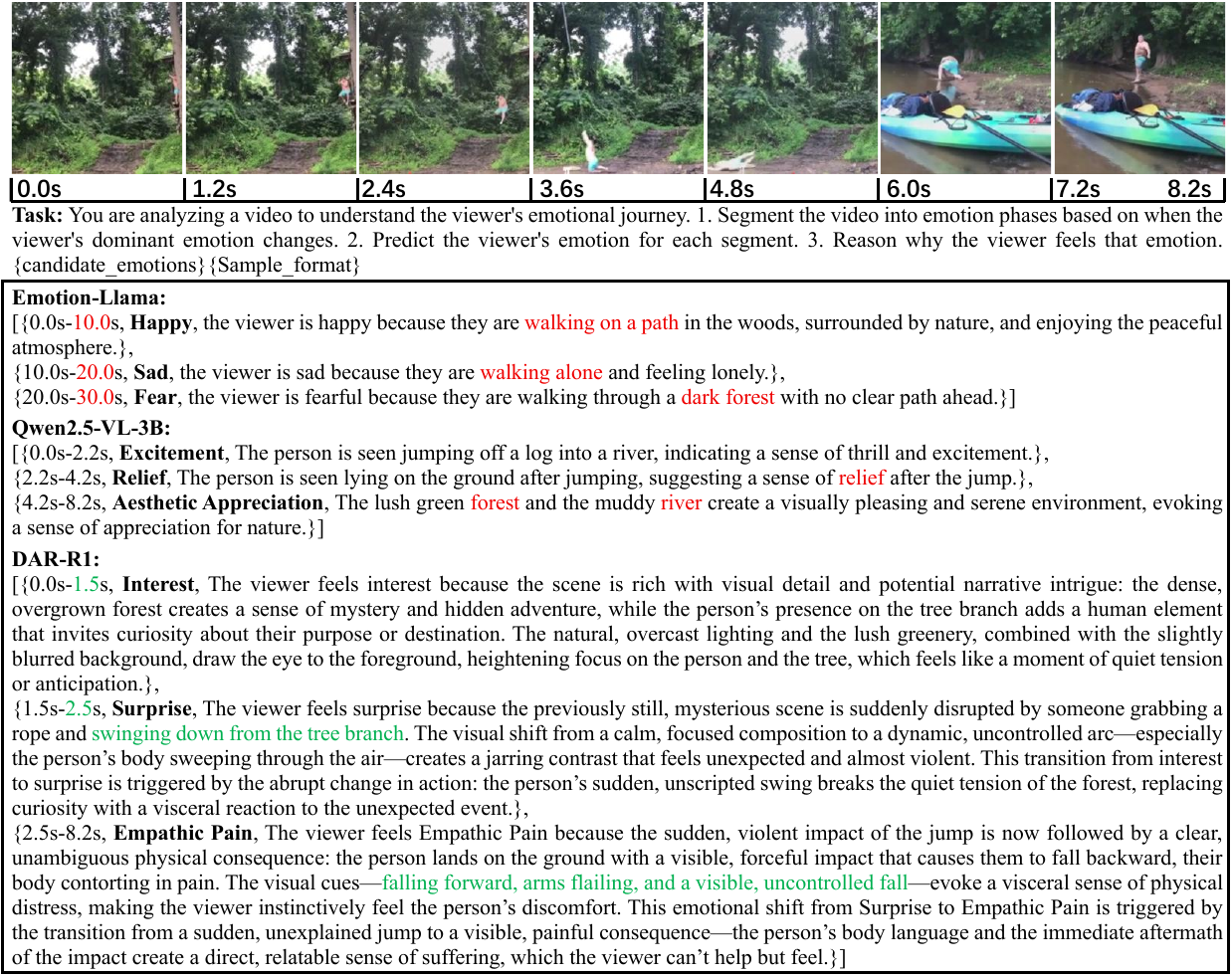}
    \caption{\textbf{Qualitative Comparison on DAR.} We visualize the temporal alignment, predicted emotion labels, and reasoning text. Green highlights indicate correct reasoning, while red indicates hallucinations or errors. DAR-R1 demonstrates superior grounding capabilities compared to baselines.}
    \label{fig:case_study}
\end{figure}

\subsection{Qualitative Analysis}
\label{subsec:qualitative}

To provide concrete insights into the capabilities of \textsc{DAR-R1}, we present a qualitative comparison in Figure~\ref{fig:case_study}.
As shown in Figure~\ref{fig:case_study}, emotion-focused MLLM fails on this benchmark. It often produces temporally invalid outputs (e.g., timestamps exceeding the video duration) and relies on generic, template-like rationales that ignore salient high-arousal events. This behavior suggests that static or coarse-grained supervision is insufficient for temporally grounded affective reasoning in videos.
The generalist model, Qwen2.5-VL-3B, correctly perceives the ``jumping'' action but fails in affective alignment. It misinterprets the crash scene—where the man hits the mud—as ``Relief'' or ``Aesthetic Appreciation'' of the lush forest. It focuses on the background scenery rather than the salient event (the accident) that drives the viewer's emotion. 
In contrast, our DAR-R1 model demonstrates precise temporal grounding and causal logic. It successfully identifies the narrative turning point. It segments the initial buildup as ``Interest'' and ``Surprise'', and crucially, it aligns the onset of ``Empathic Pain'' (2.5s) with the moment the swing loses control. Unlike Qwen2.5-VL, DAR-R1 explicitly captures the visual evidence of distress, referencing specific physical cues as highlighted in green. This causal link between the visual impact and the viewer's visceral reaction confirms that our RL training effectively aligns the model with fine-grained human emotional dynamics.

\section{Conclusion}
\label{sec:conclusion}

In this paper, we presented a comprehensive framework for transitioning Video Emotion Analysis from static classification to Dynamic Affective Reasoning. Grounded in Affective Events Theory, we introduced DAR, the first large-scale dataset featuring event-aligned segmentation and causal reasoning chains. To benchmark this task, we proposed DAR-R1, a novel training pipeline that leverages Reinforcement Learning to align MLLMs with fine-grained emotional dynamics. Extensive experiments demonstrate that our method establishes a new state-of-the-art, significantly outperforming existing models in temporal localization, emotion recognition, and reasoning quality. We hope this work serves as a foundation for advancing affective computing by enabling models to predict \textit{when} emotions shift, \textit{what} they are, and \textit{why} they arise from temporally grounded evidence.

\section*{Acknowledgements}
This work was supported by the National Science and Technology Major Project (Grant No. 2025ZD0215301), the National Natural Science Foundation of China (Grant Nos. 62402471, U22A2094, and U25A20530), the Yangtze River Delta Science and Technology Innovation Community Joint Research (Basic Research) Project (Grant No. 2025CSJZN01600), the New Generation Artificial Intelligence National Science and Technology Major Project (Grant No. 2025ZD0123100), and the Key Science and Technology Project of Anhui Province (Grant No. 202523j08050001). This research was also supported by the advanced computing resources provided by the Supercomputing Center of the University of Science and Technology of China.

\bibliographystyle{splncs04}
\bibliography{main}

\clearpage
\appendix

\section{Additional Related Work}
\subsection{Dynamic Emotion Understanding}

Dynamic emotion understanding treats affect as a time-evolving process rather than a static clip-level label. Early datasets such as Aff-Wild2~\cite{kollias2018aff} and AFEW-VA~\cite{kossaifi2017afewva} enabled dense temporal affect estimation by providing frame-level valence--arousal annotations. Subsequent works extended continuous affect tracking to multimodal interaction scenarios~\cite{ringeval2013introducing,kossaifi2019sewa} or emphasized gradual emotional transitions across utterances~\cite{barros2018omg}. More recently, TSL-300~\cite{zhang2022temporal} shifted the focus toward temporal sentiment localization in untrimmed videos. However, these datasets primarily support continuous affect regression, gradual transition modeling, or coarse sentiment localization. They do not jointly model fine-grained emotion categories, event-aligned emotional shifts, and their underlying causal mechanisms. In contrast, \textsc{DAR} requires models to identify \emph{when} fine-grained emotions shift, \emph{what} emotional states occur, and \emph{why} visual stimuli trigger these transitions.

\section{Data Reliability and Human Validation}
\label{app:data_reliability}

\begin{table}[h]
\centering
\caption{Human validation statistics on DAR annotations.}
\label{tab:human_validation}
\resizebox{0.5\linewidth}{!}{
\begin{tabular}{lc}
\toprule
Error Type & Error Rate \\
\midrule
Boundary error $>$0.5s & 2.4\% \\
Emotion-label disagreement & 3.1\% \\
Visual-grounding error & 2.2\% \\
Non-viewer-centric rationale & 1.3\% \\
\bottomrule
\end{tabular}}
\end{table}

Beyond the three-stage verification in the annotation pipeline, DAR annotations are further validated by three experts on the full test split and a random subset of training samples. The validation covers boundary accuracy, emotion-label consistency, visual grounding, and viewer-centric rationales. The validation statistics are summarized in Table~\ref{tab:human_validation}.

\section{Visual Diversity Analysis}
\label{app:visual_diversity}

We conduct a post-hoc visual-diversity analysis based on the segment-level descriptions $D_t$ in DAR. The genre distribution is diverse: 27.2\% Nature/Wildlife, 20.5\% Animals, 12.2\% Comedy, 8.9\% Urban Life, 8.0\% Horror, 7.3\% Food/Cooking, 7.2\% Sports, 6.9\% Performance, and 1.8\% Others. The scene distribution is also balanced, with 52.0\% outdoor scenes and 48.0\% indoor scenes. These results indicate that DAR covers rich visual scenarios and diverse scene contexts.

\section{Dataset Split}
\label{app:dataset_split}

To preserve the distribution of dynamic affective patterns, the split is stratified by dominant emotion, number of segments, video duration, and coarse visual genre/scene type. 
This results in 13,646 training videos and 1,441 test videos. 
Hyperparameters and model selection are tuned only on a held-out subset of the training split, while the test split is used exclusively for final evaluation.

\section{Additional Fine-Tuned Emotion-Focused Baselines}
\label{app:fair_comparison}

\begin{table}[h]
\centering
\caption{Fine-tuned emotion-focused baselines on DAR.}
\label{tab:app_fair_comparison}
\resizebox{0.8\linewidth}{!}{
\begin{tabular}{lccccc}
\toprule
Model & SC-Acc & mIoU & Emo-Acc & ROUGE-L & SBERT \\
\midrule
Videmo-FT & 10.3 & 22.1 & 5.0 & 14.7 & 50.2 \\
EmotionLlama-FT & 31.4 & 42.9 & 5.4 & 17.5 & 54.6 \\
AffectGPT-FT & 36.2 & 46.5 & 6.6 & 20.2 & 56.3 \\
StimuVAR-FT & 13.5 & 16.3 & 10.2 & 23.9 & 61.1 \\
DAR-SFT & 37.7 & 47.6 & 25.8 & 24.1 & 77.0 \\
\textbf{DAR-R1} & \textbf{41.5} & \textbf{52.3} & \textbf{28.6} & \textbf{24.8} & \textbf{78.6} \\
\bottomrule
\end{tabular}}
\end{table}

To address the concern that existing emotion-focused MLLMs are not originally designed for DAR, we additionally fine-tune four emotion-focused MLLMs on DAR. As shown in Table~\ref{tab:app_fair_comparison}, DAR-R1 still outperforms these stronger baselines.

\section{Out-of-Domain Evaluation}
\label{app:ood}

\begin{table}[h]
\centering
\caption{OOD temporal localization results (\%) on TSL.}
\label{tab:app_ood_tsl}
\resizebox{0.8\linewidth}{!}{
\begin{tabular}{lccccc}
\toprule
Model & mAP@0.1 & mAP@0.15 & mAP@0.2 & mAP@0.25 & mAP@0.3 \\
\midrule
Videmo~\cite{zhang2025videmo} & 4.1 & 2.7 & 1.7 & 1.2 & 1.0 \\
EmotionLlama~\cite{cheng2024emotion} & 4.8 & 3.0 & 1.9 & 1.3 & 1.1 \\
AffectGPT~\cite{lian2025affectgpt} & 7.9 & 6.1 & 4.8 & 3.9 & 3.1 \\
StimuVAR~\cite{guo2025stimuvar} & 2.7 & 1.5 & 0.6 & 0.3 & 0.3 \\
\textbf{DAR-R1} & \textbf{13.2} & \textbf{11.4} & \textbf{9.4} & \textbf{7.5} & \textbf{6.7} \\
\bottomrule
\end{tabular}}
\end{table}

We further conduct an out-of-domain evaluation on TSL~\cite{zhang2022temporal}, following the temporal localization protocol. As shown in Table~\ref{tab:app_ood_tsl}, DAR-R1 remains consistently stronger than emotion-focused baselines, indicating that the proposed dynamic formulation and reward-guided training generalize beyond the in-domain DAR split.

\section{Reward Ablation Studies}
\label{app:reward_ablation}

\begin{table}[h]
\centering
\caption{Ablation study of task rewards on DAR.}
\label{tab:app_reward_ablation}
\resizebox{0.7\linewidth}{!}{
\begin{tabular}{lccccc}
\toprule
Variant & SC-Acc & mIoU & Emo-Acc & ROUGE-L & SBERT \\
\midrule
w/o $R_{\rm count}$ & 38.3 & 49.7 & 26.5 & 24.2 & 77.4 \\
w/o $R_{\rm seg}$ & 39.4 & 48.0 & 27.0 & 24.5 & 77.7 \\
w/o $R_{\rm emo}$ & 40.0 & 51.0 & 24.2 & 24.6 & 78.2 \\
\textbf{DAR-R1} & \textbf{41.5} & \textbf{52.3} & \textbf{28.6} & \textbf{24.8} & \textbf{78.6} \\
\bottomrule
\end{tabular}}
\end{table}

Table~\ref{tab:app_reward_ablation} reports ablation studies on the main task rewards. Removing each reward leads to a performance decline in its corresponding dimension, confirming the necessity of the reward design. We keep $R_{\rm struct}$ and $R_{\rm reason}$ as stability rewards for GRPO training: $R_{\rm struct}$ enforces valid JSON, monotonic timestamps, and non-fragmented segments, while $R_{\rm reason}$ controls rationale length and repetition. Without $R_{\rm reason}$, the model tends to generate excessively long rationales that exceed the token budget and destabilize GRPO training.

\section{Qualitative Inference Examples of DAR-R1}
\label{app:dar_r1_inference_examples}

\begin{figure*}[h]
    \centering
    \includegraphics[width=0.75\linewidth]{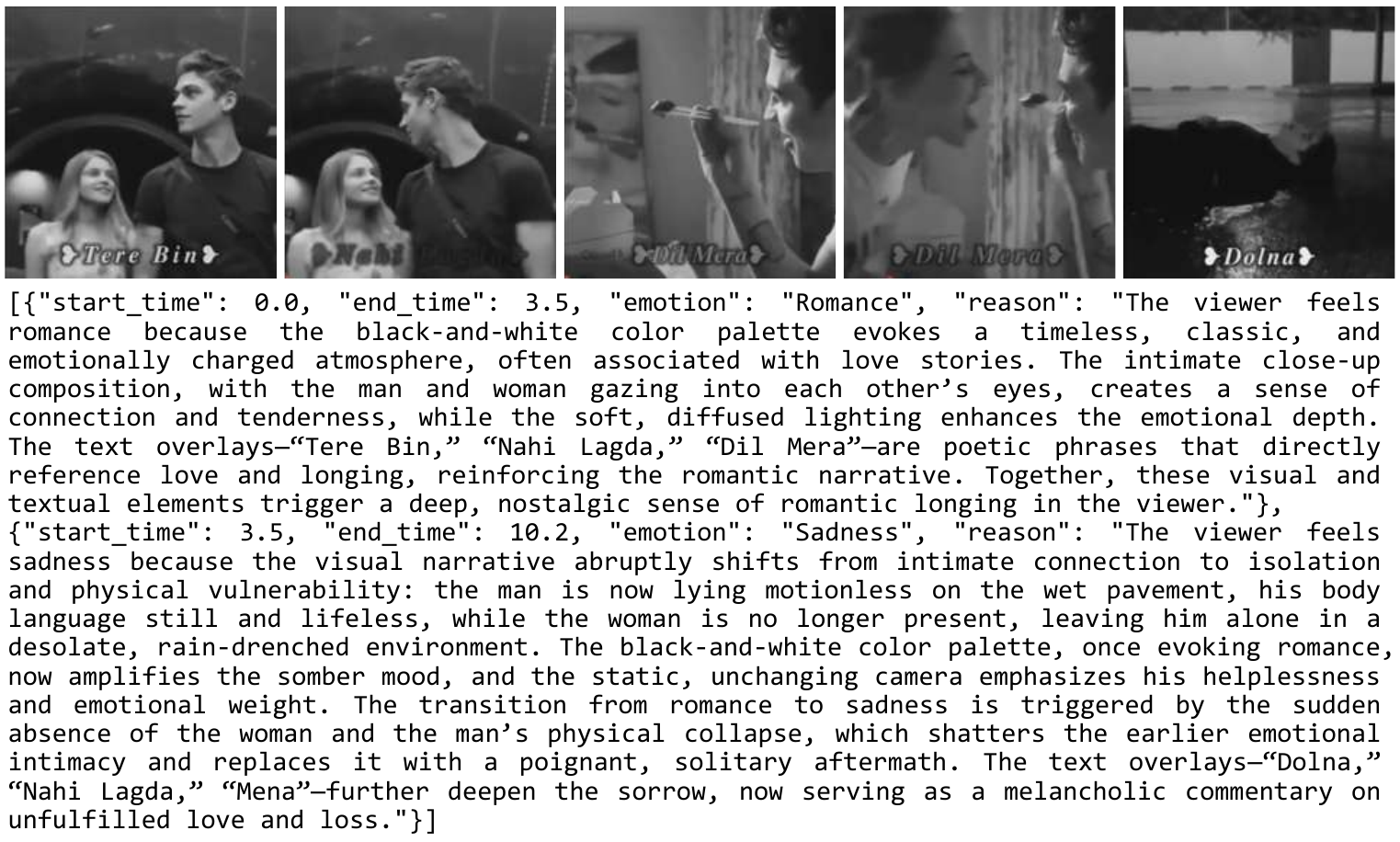}
    \caption{Qualitative inference example of DAR-R1.}
    \label{fig:dar_r1_inference_example}
\end{figure*}

We provide a qualitative inference example of DAR-R1 in Fig.~\ref{fig:dar_r1_inference_example}. 

\section{LLM-as-a-Judge Prompt}
\label{app:judge_prompt}

We evaluate the quality of the predicted \texttt{emotion} and \texttt{reason} pairs using an LLM-as-a-judge protocol. 
For each predicted segment, the judge receives the video, the target temporal interval, the task question, and the model prediction in JSON format. 

\paragraph{LLM-as-a-judge prompt.}
We use the following prompt template:

\begingroup
\footnotesize
\begin{verbatim}
You are an expert evaluator for viewer-centric dynamic affective reasoning 
in videos.
You will be given:
1. A silent video.
2. A target temporal segment: [START_TIME, END_TIME].
3. A question: "What emotion would a viewer feel in this segment, and why?"
4. A model prediction containing: predicted emotion and reason.
Important:
- The emotion refers to the VIEWER's induced emotion, not the emotion 
expressed by people in the video.
- The reason should be grounded in visible evidence.
- The answer should focus on the target temporal segment.
- If the segment is not the first segment, consider whether the predicted 
emotion and reason are temporally consistent with the preceding emotional 
context.
Target segment:
[START_TIME, END_TIME]
Model prediction:
Emotion: [PREDICTED_EMOTION]
Reason: [PREDICTED_REASON]
Evaluate the predicted emotion--reason pair from 0 to 5.
Score 5: The answer is fully correct and well grounded.
Score 4: The answer is Mostly correct and largely grounded.
Score 3: The answer is partially correct. 
Score 2: The answer has limited correctness. 
Score 1: The answer is mostly incorrect. 
Score 0: The answer is invalid or irrelevant.
Evaluate the answer along the following dimensions:
- visual_grounding: Is the reason supported by visible content in the 
segment?
- causal_logic: Does the reason explain why the visual event induces the
  viewer emotion?
- viewer_centricity: Does it describe the viewer's feeling rather than the
  character's emotion?
- temporal_consistency: Is the reason grounded in the specified time 
interval, consistent with the emotional transition, and coherent with the 
preceding segment context when applicable?
- answer_consistency: Are the emotion label and reason mutually consistent?
Output only a valid JSON object in the following format:
{ "score": <integer from 0 to 5>,
  "visual_grounding": <integer from 0 to 5>,
  "causal_logic": <integer from 0 to 5>,
  "viewer_centricity": <integer from 0 to 5>,
  "temporal_consistency": <integer from 0 to 5>,
  "answer_consistency": <integer from 0 to 5>,
  "brief_comment": "<one-sentence explanation>"}
\end{verbatim}
\endgroup

\section{Examples of DAR Data Construction}
\label{app:data_construction_examples}
\begin{figure*}[t]
    \centering
    \includegraphics[width=0.95\linewidth]{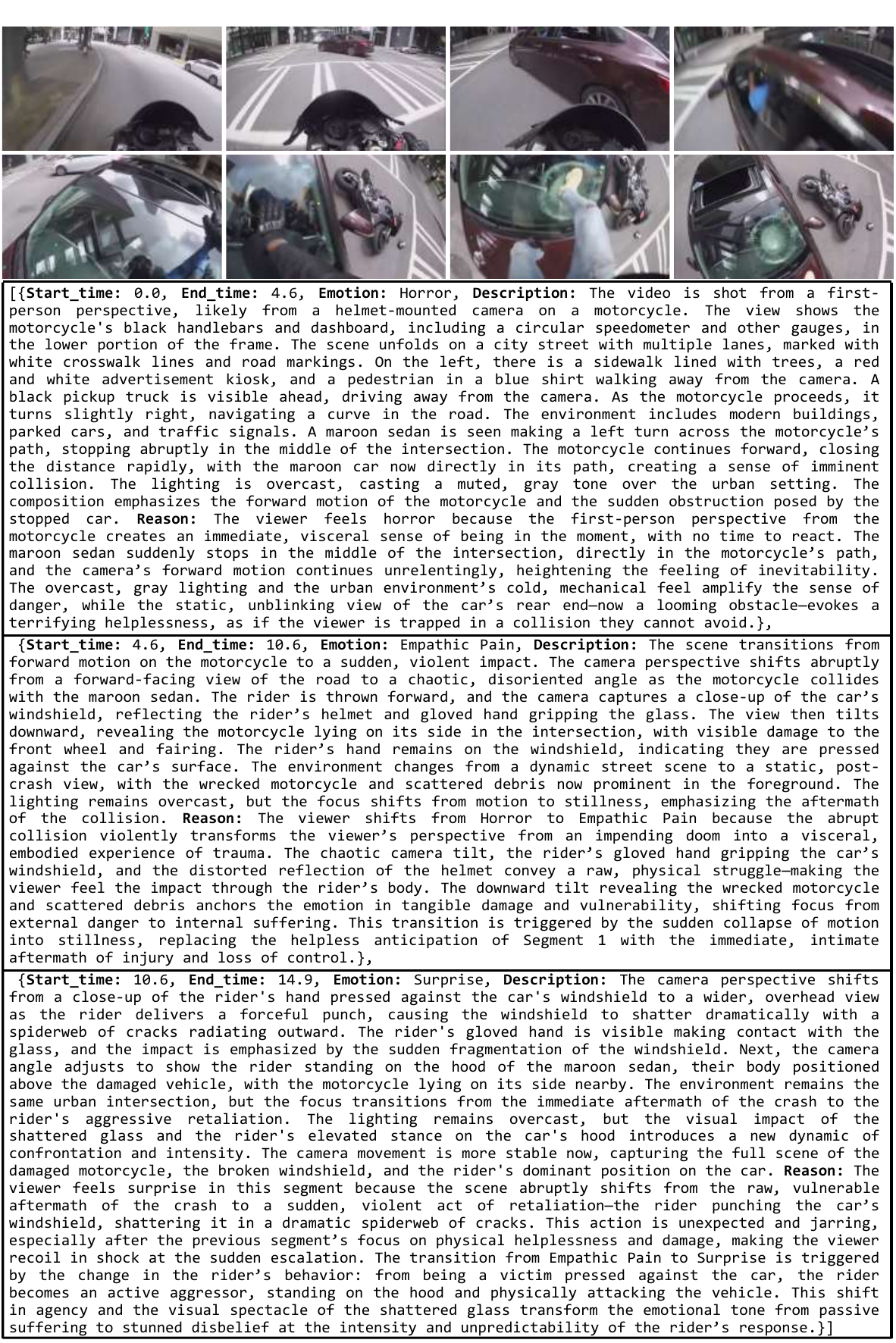}
    \caption{Example of the DAR data construction process.}
    \label{fig:dar_construction_example}
\end{figure*}

We provide an example of the DAR construction process in Fig.~\ref{fig:dar_construction_example}. 

\end{document}